\def\year{2020}\relax
\documentclass[letterpaper]{article}
\usepackage{aaai20}
\usepackage{times}
\usepackage{helvet}
\usepackage{courier}
\usepackage{url}
\usepackage{graphicx}
\usepackage{amsfonts,amsmath}
\usepackage{booktabs,multirow,array,color}
\title{Joint Learning of Answer Selection and Answer Summary Generation in Community Question Answering\thanks{This work was financially supported by the National Natural Science Foundation of China (No.61602013) and a grant from the Research Grant Council of the Hong Kong Special Administrative Region, China (Project Code: 14204418).}}
\author{
Yang Deng\textsuperscript{\rm 1},
Wai Lam\textsuperscript{\rm 1},
Yuexiang Xie\textsuperscript{\rm 3},
Daoyuan Chen\textsuperscript{\rm 4},
Yaliang Li\textsuperscript{\rm 4},
Min Yang\textsuperscript{\rm 5},
Ying Shen\textsuperscript{\rm 23$\dagger$}\\
\textsuperscript{\rm $\dagger$}Corresponding Author\\
\textsuperscript{\rm 1}The Chinese University of Hong Kong,
\textsuperscript{\rm 2}South China University of Technology,\\
\textsuperscript{\rm 3}Peking University Shenzhen Graduate School,
\textsuperscript{\rm 4}Alibaba Group,
\textsuperscript{\rm 5}Chinese Academy of Sciences,\\
{\tt \{ydeng,wlam\}@se.cuhk.edu.hk, \{xieyx,shenying\}@pku.edu.cn,}\\
  {\tt \{daoyuanchen.cdy,yaliang.li\}@alibaba-inc.com, min.yang@siat.ac.cn}
}

\begin{document}
\maketitle
\begin{abstract}
Community question answering (CQA) gains increasing popularity in both academy and industry recently. However, the redundancy and lengthiness issues of crowdsourced answers limit the performance of answer selection and lead to reading difficulties and misunderstandings for community users. To solve these problems, we tackle the tasks of answer selection and answer summary generation in CQA with a novel joint learning model. Specifically, we design a question-driven pointer-generator network, which exploits the correlation information between question-answer pairs to aid in attending the essential information when generating answer summaries. Meanwhile, we leverage the answer summaries to alleviate noise in original lengthy answers when ranking the relevancy degrees of question-answer pairs. In addition, we construct a new large-scale CQA corpus, WikiHowQA, which contains long answers for answer selection as well as reference summaries for answer summarization. The experimental results show that the joint learning method can effectively address the answer redundancy issue in CQA and achieves state-of-the-art results on both answer selection and text summarization tasks. Furthermore, the proposed model is shown to be of great transferring ability and applicability for resource-poor CQA tasks, which lack of reference answer summaries.
\end{abstract}

\section{Introduction}
Recent years have witnessed a spectacular increase in real-world applications of community question answering (CQA), such as Yahoo! Answer\footnote{https://answers.yahoo.com/} and StackExchange\footnote{https://stackexchange.com/}. Many studies have been made on different tasks in CQA, such as answer selection, question-question relatedness, and comment classification~\cite{DBLP:conf/eacl/MoschittiBU17,DBLP:conf/emnlp/JotyMN18,DBLP:conf/semeval/NakovHMMMBV17}. However, due to the length and redundancy of answers in CQA scenario, there are several challenges that need to be tackled in real-world applications. (i) The noise introduced by the redundancy of answers makes it difficult for answer selection model to pick out correct answers from a set of candidates. (ii) Compared with other QA systems (e.g., factoid question answering), answers in CQA are often too long for community users to read and comprehend.

Current state-of-the-art answer selection models~\cite{Tan2016Improved,DBLP:conf/acl/WuWS18} employ the attention mechanism to attend the important correlated information between question-answer pairs. These methods perform well when ranking short answers, while the accuracy goes down with the increase in the length of answers~\cite{dos2016attentive,COALA}.
Recent studies on coarse-to-fine question answering for long documents, such as Reading Comprehension (RC)~\cite{DBLP:conf/acl/ChoiHUPLB17,DBLP:conf/aaai/WangYGWKZCTZJ18,DBLP:conf/acl/XiaoWLWL18}, focus on the answer span extraction in factoid QA, in which those factoid questions can be answered by a certain word or a short phrase. Conversely, in non-factoid CQA, discrete and complex information from multiple sentences makes up the answers together. Besides, generative RC methods~\cite{DBLP:conf/acl/NishidaSNSOAT19} only give one certain answer, while there are often multiple useful answers in CQA. Thus, these approaches are not suitable for addressing the redundancy issue of answers in CQA.

On the other hand, text summarization provides an effective approach to alleviating the aforementioned issue. Text summarization methods can generally be divided into two categories: extractive summarization~\cite{DBLP:conf/acl/0001L16,DBLP:conf/aaai/NallapatiZZ17} and abstractive summarization~\cite{DBLP:conf/acl/SeeLM17,DBLP:conf/conll/NallapatiZSGX16}. The aim is to assemble or generate summaries from the source article or external vocabulary, based on the information from the source text. In the existing studies, answer summarization in CQA is mainly explored by extractive summarization models~\cite{DBLP:conf/acl/TomasoniH10,DBLP:conf/wsdm/SongRLLMR17}. However, due to the length of answers, extractive methods sometimes fall short of generalization of all the important information in the whole answer and consistency of the core idea.  Besides, the correlation information between question and answer, which plays a crucial role in human comprehension, is underutilized by current query-based summarization studies~\cite{DBLP:conf/acl/NemaKLR17,DBLP:conf/ecir/SinghMOBK18}. Therefore, we intend to take advantage of both the contextual information from the source text and the relationship between the question-answer pair to generate abstractive answer summaries in CQA.

We aim to simultaneously tackle the above issues in CQA, including (i) improving the performance of non-factoid answer selection with long answers, (ii) generating abstractive summaries of the answers. We jointly learn answer selection and abstractive summarization to generate answer summaries for CQA. First, we exploit the correlated information between question-answer pairs to improve abstractive answer summarization, which enables the summarizer to generate abstractive summaries related to questions. Then, we measure the relevancy degrees between questions and answer summaries to alleviate the impact of noise from original answers. Besides, since obtaining reference summaries is usually labor-intensive and time-consuming in a new domain, a transfer learning strategy is designed to improve resource-poor CQA tasks with large-scale supervision data.

We summarize our contributions as follows:

1. We jointly learn answer selection and answer summary generation to tackle the lengthiness and redundancy issues of the answer in CQA with a unified model. A novel joint learning framework of answer selection and abstractive summarization (ASAS) is proposed to employ the question information to guide the abstractive summarization, and meanwhile leverage the summaries to reduce noise in answers for precisely measuring the correlation degrees of QA pairs.

2. We construct a new dataset, WikiHowQA, for the task of answer summary generation in CQA, which can be adapted to both answer selection and summarization tasks. Experimental results on WikiHowQA show that the proposed joint learning method outperforms SOTA answer selection methods and meanwhile generates more precise answer summaries than existing summarization methods. 

3. To handle resource-poor CQA tasks, we design a transfer learning strategy, which enable those tasks without reference answer summaries to conduct the joint learning with impressive experimental results. 

\section{Related Work}

\noindent\textbf{Community Question Answering.} 
Answer selection is the core and the most widely-studied problem in community question answering. Recent studies have evolved from feature-based methods~\cite{DBLP:conf/sigir/WangMC09,DBLP:conf/coling/WangM10a} into deep learning models, such as convolutional neural network (CNN)~\cite{Severyn2015Learning} and recurrent neural network (RNN)~\cite{DBLP:conf/acl/WangN15}. In order to capture the interactive information in QA sentences, various attention mechanisms~\cite{Tan2016Improved,dos2016attentive} are developed to align the related words between questions and answers. However, the lengthy and redundant answers in CQA scenario may introduce much noise and scatter important information, which causes difficulties in answer selection. Some studies leverage additional information to compensate the imbalance of information between questions and answers, such as user model~\cite{DBLP:conf/aaai/WenMFZ18,DBLP:conf/wsdm/LiDLXFLGS17}, latent topic~\cite{DBLP:conf/naacl/YoonSJ18}, external knowledge~\cite{DBLP:conf/sigir/ShenDYLD0L18} or question subject~\cite{DBLP:conf/acl/WuWS18}. Some existing transfer learning studies on CQA focus on cross-domain adaptation~\cite{DBLP:conf/coling/DengSYLDFL18,DBLP:conf/wsdm/YuQJHSCC18}. In this work, we employ summarization method to reduce noise in the original lengthy answers to improve the answer selection performance in CQA.

\noindent\textbf{Text Summarization.} Text summarization techniques are mainly classified into two categories: extractive and abstractive summarization. Extractive approaches regard summarization as a sentence classification~\cite{DBLP:conf/aaai/NallapatiZZ17} or a sequence labeling task~\cite{DBLP:conf/acl/0001L16} to select sentences from the article to form the summary, while abstractive approaches usually employ attention-based encoder-decoder models~\cite{DBLP:conf/conll/NallapatiZSGX16,DBLP:conf/acl/SeeLM17} to generate abstractive summaries. Answer summarization in CQA was first introduced by~\citeauthor{DBLP:conf/lrec/ZhouLH06} (2006) as an application of extractive summarization. After that, studies on answer summarization are still regarded as a separate extractive summarization module in QA pipeline~\cite{DBLP:conf/acl/TomasoniH10,DBLP:conf/wsdm/SongRLLMR17}. Besides, query-based summarization methods~\cite{DBLP:conf/acl/NemaKLR17,DBLP:conf/ecir/SinghMOBK18} also can be a good solution for this task, however, these approaches are reported to perform worse than answer selection methods on question answering scenario~\cite{DBLP:conf/acl/KhapraSSA18}.

\noindent\textbf{Multi-task Learning.} 
Inspired by the success of multi-task learning in other NLP tasks, several attempts have been made to solve answer selection with different tasks. \citeauthor{DBLP:conf/eacl/MoschittiBU17} (2017) and \citeauthor{DBLP:conf/emnlp/JotyMN18} (2018) enhance answer selection in CQA via multi-task learning with the auxiliary tasks of question-question relatedness and question-comment relatedness. \citeauthor{DBLP:conf/ijcai/0007CCWZS19} (2019) leverage the question categorization to
enhance the question representation learning for CQA. \citeauthor{DBLP:conf/aaai/DengXLYDFLS19} (2019) propose a multi-view attention based multi-task learning model to jointly tackle answer selection and knowledge base question answering tasks. In this work, we jointly learn answer selection and abstractive summarization to select and generate precise answers in CQA.

\section{Method}

\subsection{Problem Definition}
We aim to jointly conduct two tasks, answer selection and abstractive summarization, to select and generate concise answers for CQA. Given a question $q_i$, the goal is to simultaneously select the set of correct answers from a set of candidates $A_i=\{a^{(1)}_i,...,a^{(j)}_i\}$ and generate an abstractive summary $\beta^{(*)}_i$ for each selected answer $a^{(*)}_i$. 

The dataset $D$ for learning typically contains a set of questions $Q$ with the number of $N$. For each question $q_i \in Q$, there are $M_i$ candidate answers $A_i$ with the corresponding reference summary $\beta^{(j)}_i$ written by human and the label $y^{(j)}_i$ determining whether $a^{(j)}_i$ can answer $q_i$. 
\begin{equation}
 D = \{(q_i,\{(a^{(j)}_i,\beta^{(j)}_i,y^{(j)}_i)\}^{M_i}_{j=1})\}^{N}_{i=1}.
\end{equation}

\subsection{Model}
\begin{figure}
\centering
\includegraphics[width=0.5\textwidth]{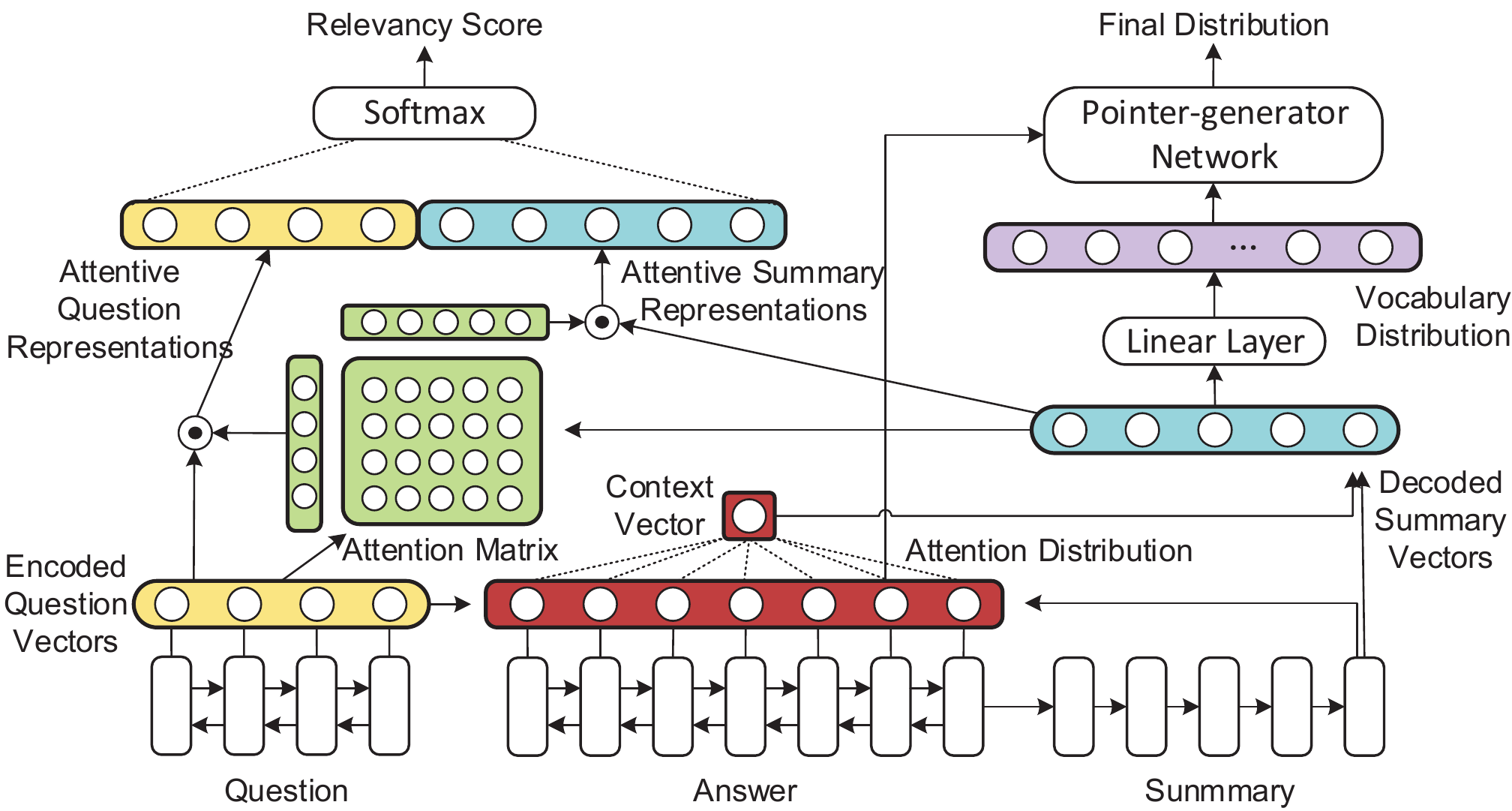}
\caption{The Joint Learning Framework of Answer Selection and Abstractive Summarization (ASAS).}
\label{method}
\end{figure}
We introduce the proposed joint learning model for answer selection and abstractive summarization (ASAS). As is depicted in Fig.~\ref{method}, The overall framework of ASAS consists of four components: (i) Shared Compare-Aggregate Bi-LSTM Encoder, (ii) Sequence-to-sequence Model with Question-aware Attention, (iii) Question Answer Alignment with Summary Representations, (iv) Question-driven Pointer-generator Network. 

\subsubsection{Shared Compare-Aggregate Bi-LSTM Encoder.}
The word embeddings of the question and the original answer, $W_q$ and $W_a$, are fed into a compare layer as~\citeauthor{compare-aggregate} (2017) to generate the model input $\hat{W_q}$ and $\hat{W_a}$. Then, a pair of Bi-LSTM encoders are adopted to aggregate the context information. 
We encode a pair of word sequences of the question $q$ and the answer $a$ into sentence representations $H\in\mathbb{R}^{L\times{d_h}}$, where $L$ and $d_h$ are the length of sentences and the size of hidden states:
\begin{gather}
H_{q} = \textbf{Bi-LSTM}(\hat{W_q}), \quad H_{a} = \textbf{Bi-LSTM}(\hat{W_a}).
\end{gather}

\subsubsection{Seq2Seq Model with Question-aware Attention.}
With the intuition that the information in the question is supposed to be helpful in attending the important elements in the original answer sentence, we propose a question-aware attention based seq2seq model to decode the encoded sentence representation of the answer. We adopt a unidirectional LSTM as the decoder. On each step $t$, the decoder produces the hidden state $s_t$ with the input of the previous word $w_{t-1}$. The question-aware attention $\alpha^t$ is generated by:
\begin{gather}
s_t= \textbf{LSTM}(s_{t-1},w_{t-1}),\\
o_q = Average(H_q);\\
e_i^t = m^t \text{tanh} (W_h h^a_i+W_s s_t+W_q o_q+b), \\
\alpha^t = \text{softmax}(e^t),
\end{gather}
where $m$, $W_h$, $W_s$, $W_q$ are attention parameter matrices to be learned. The question-aware attention weight $\alpha^t$ is used to generate context vector $\hat{h}_t$ as a probability distribution over the source words:
\begin{equation}
    \hat{h}_t = \sum\nolimits_i \alpha^t h^a_i.
\end{equation}

The context vector aggregates the information from the source text and the question for the current step. We concatenate it with the decoder state $s_t$ and pass through a linear layer to generate the summary representation $h^s_t$:
\begin{equation}
    h^s_t=W_1[s_t:\hat{h}_t]+b_1,
\end{equation}
where $W_1$ and $b_1$ are parameters to be learned.

\subsubsection{Question Answer Alignment with Summary Representations.}
We apply a two-way attention mechanism to generate the co-attention between the encoded question representation $H_q$ and the decoded summary representation $H_s$:
\begin{gather}
 M_{qa}=\text{tanh}\left(H_q^TUH_s\right),\\
 \alpha_q = \text{softmax}(\text{Max}(M_{qa})),\\
 \alpha_a =\text{softmax}(\text{Max}({M_{qa}}^T)),
\end{gather}
where $U\in\mathbb{R}^{d_{s}\times{d_{s}}}$ is the attention parameter matrix to be learned; $d_{s}$ is the dimension of QA representations; $\alpha_q$ and $\alpha_a$ are the co-attention weights for the question and the answer summary respectively.

We conduct dot product between the attention vectors and the question and summary representations to generate the final attentive sentence representations for answer selection:
\begin{align}
 r_q = H_q^T\alpha_q, \quad r_a = H_s^T\alpha_a.
\end{align}

Compared with encoded answer representations, decoded summary representations are more concise and compressive, which enable answer selection model to precisely capture the interactive information between questions and answers.

\subsubsection{Question-driven Pointer-generator Network.}
First, the probability distribution $P_{vocab}$ over the fixed vocabulary is obtained by passing the summary representation $h^s_t$ through a softmax layer:
\begin{equation}
    P_{vocab} = \text{softmax}(W_2 h^s_t + b_2),
\end{equation}
where $W_2$ and $b_2$ are parameters to be learned. Then, a question-aware pointer network is proposed to copy words from the source article with the guidance of the question information. The question-aware generation probability $p_{gen}\in [0,1]$ takes into account the decoded summary representation $h^s_t$, the decoder input $x_t$ and the question representation $o_q$:
\begin{equation}
    p_{gen} = \sigma (w_h^T h^s_t + w_x^T x_t + w_q^T o_q + b_p),
\end{equation}
where $w_h$, $w_x$, $w_q$ and $b_p$ are parameters to be learned, and $\sigma$ is the sigmoid function. Following the basic pointer-generator network (PGN)~\cite{DBLP:conf/acl/SeeLM17}, we obtain the final probability distribution over both the fixed vocabulary and words from the source article:
\begin{equation}
    P = p_{gen}P_{vocab} + (1-p_{gen})\sum\nolimits_{i:w_i=w} \alpha^t_i.
\end{equation}

To be specific, the question information is involved in not only the generating process, but also the copying process in the question-driven PGN. (i) The question information directs the calculation of the generation probability to decide whether generating a word from the vocabulary or copying from the source text. (ii) The question-aware attention weights integrate the question information to attend the important words in the source text for copying. (iii) The probability distribution over the vocabulary is learned from the question-aware attentive summary representations.

\subsection{Joint Training Procedure}
\subsubsection{Answer Selection Loss.}
The attentive representations of questions and summaries go through a softmax layer for binary classification: 
\begin{equation}
y(q,a)=\text{softmax}\left(W_s[r_q:r_a]+b_s\right),
\end{equation}
where $W_s\in\mathbb{R}^{d_x\times{2}}$ and $b_s\in\mathbb{R}^2$ are  parameters to be learned. The answer selection task is trained to minimize the cross-entropy loss function:
\begin{equation}
 L_{qa}=-\sum_{i=1}^N\left[y_i\log{p_i}+\left(1-y_i\right)\log{\left(1-p_i\right)}\right],
\end{equation}
where $p$ is the output of the softmax layer and $y$ is the binary classification label of the QA pair.

\subsubsection{Summarization Loss.}
The summarization task is trained to minimize the negative log likelihood:
\begin{equation}
    L_{sum} = - \frac{1}{T}\sum^T_{t=0}\text{log}P(w_t^*).
\end{equation}

\subsubsection{Coverage Loss.}
Coverage loss~\cite{DBLP:conf/acl/SeeLM17} was proposed to discourage the repetition in abstractive summarization. In each decoder timestep $t$, the coverage vector $c^t=\sum^{t-1}_{t'=0}a^{t'}$ is used to represent the degree of coverage so far. The coverage vector $c^t$ will be applied to compute the attention weight $\alpha^t$. The coverage loss is trained to penalize the repetition in updated attention weight $\alpha^t$:
\begin{equation}
    L_{cov}=\frac{1}{T}\sum^T_{t=1}\sum\nolimits_i \text{min}(\alpha^t_i,c^t_i).
\end{equation}

\subsubsection{Overall Loss Function.}
For joint training, the final objective function is to minimize above three loss functions:
\begin{equation}
    L=\lambda_1 L_{qa}+\lambda_2 L_{sum}+\lambda_3 L_{cov},
\end{equation}
where $\lambda_1$, $\lambda_2$, $\lambda_3$ are hyper-parameters to balance losses.

\subsection{Handling Resource-poor Datasets}
Since annotating gold answer summaries is a labor-intensive work, we intend to leverage the knowledge learned from the joint learning of answer selection and answer summary generation on a large-scale supervision dataset and apply it to resource-poor datasets without reference answer summaries. 
The goal can be achieved by a transfer learning strategy involving two steps: (i) initialize the the parameters of model pre-trained on the source dataset, (ii) further fine-tune on the target dataset. A straightforward way is to fine-tune all the parameters learned from the source data on the target training dataset. Another fashion is to fine-tune a certain part of parameters and keep the remaining part of model fixed during fine-tuning. In this case, we first pre-train the whole joint learning model on the source dataset, and then only fine-tune the answer selection modules (including Shared Compare-Aggregate Bi-LSTM Encoder \& Question Answer Alignment). On one hand, fixing the summarization part can not only reduce the demand for annotating summary data, but also prevent model over-fitting. On the other hand, questioning styles and answer contents vary from CQA tasks in different domains, thus, the answer selection part is supposed to benefit from fine-tuning in target domains.

\section{Datasets and Experimental Setting}

\begin{table}
\fontsize{8}{9}\selectfont
\centering
  \begin{tabular}{cc}
  \toprule
  & train / dev / test\\
  \midrule
  \#Questions &76,687 / 8,000 / 22,354\\
  \#QA Pairs & 904,460 / 72,474 / 211,255\\
  \#Summaries &142,063 / 18,909 / 42,624\\
  Avg QLen & 7.20 / 6.84 / 6.69\\
  Avg ALen& 520.87 / 548.26 / 554.66\\
  Avg SLen& 67.38 / 61.84 / 74.42\\
  Avg \#CandA& 11.79 / 9.06 / 9.45 \\
  \bottomrule
  \end{tabular}
\caption{Statistic of WikiHowQA Dataset}
\label{stat}
\end{table}

\subsection{Datasets}
Most of the widely-adopted answer selection benchmark datasets are composed of short sentences, such as WikiQA~\cite{Yang2015WikiQA}, SemEval~\cite{DBLP:conf/semeval/NakovHMMMBV17}. WikiPassageQA~\cite{DBLP:conf/sigir/CohenYC18} and StackExchange~\cite{COALA}, two latest non-factoid answer selection datasets with long passages (about 150 words) as candidate answers, lack of the reference summary for answer summarization evaluation in our defined answer summary generation task. 

We present a new CQA corpus, WikiHowQA, for answer summary generation, which contains labels for the answer selection task as well as reference summaries for the text summarization task. To prepare this dataset, we modify a latest text summarization dataset, WikiHow~\cite{DBLP:journals/corr/abs-1810-09305}, which was obtained from \emph{WikiHow}\footnote{http://www.wikihow.com/} knowledge base. The WikiHow dataset contains detailed answers written by community users for non-factoid questions starting with ``\textbf{How to}". The original answers are composed by multiple steps of different methods for the question, and the description in each step is associated with an abstractive summary. The WikiHow dataset only contains the selected ground-truth answers and the reference summaries for each answer, while the whole candidate answer set is required when we wish to conduct answer selection experiments on this dataset. Therefore, we construct a new CQA dataset based on the WikiHow dataset.

We first clean up the WikiHow dataset by filtering out those questions without answers or summaries and those answers with punctuation only. After that, the dataset size is reduced from 230,843 to 203,596, including 107,041 unique questions. The clean WikiHow dataset is split into 142,063 / 18,909 / 42,624 as train / dev / test sets. In order to retrieve the candidate answer pool for all the questions, we write a crawler to collect the relevant questions for each question from the \emph{WikiHow} website. 
The answers of the relevant questions posted on \emph{WikiHow} are labeled as negative answers for the given question.
Finally, we obtain 1,188,189 question-answer pairs with corresponding answer summaries and matching labels as the WikiHowQA dataset. In accordance with the clean WikiHow dataset, we split the WikiHowQA dataset into 904,460 / 72,474 / 211,255 as train / dev / test sets, which implies that there is no overlapping of samples among the three split sets. The statistics of the WikiHowQA\footnote{https://github.com/dengyang17/wikihowQA} dataset are shown in Table~\ref{stat}. 

\begin{table}
\fontsize{8}{10}\selectfont
\centering
  \begin{tabular}{ccc}
  \toprule
  \multirow{2}{*}{StackExchange} & \#Questions &\multirow{2}{*}{Avg ALen} \\
  &(train/dev/test)&\\
  \midrule
  Travel & 3,572 / 765 / 766  &214\\
  Cooking &3,692 / 791 / 792 & 189\\
  Academia & 2,856 / 612 / 612  &229\\
  Apple & 5,831 / 1,249 / 1,250 & 114\\
  Aviation & 3,035 / 650 / 652  &281\\
  \bottomrule
  \end{tabular}
\caption{Statistic of StackExchange CQA Dataset}
\label{stat3}
\end{table}

In addition, we evaluate the proposed method on a resource-poor CQA dataset, StackExchange~\cite{COALA}, which lacks of reference answer summaries. The statistics of the StackExchange dataset are presented in Table~\ref{stat3}, which is a real-life CQA dataset containing data with long answers from different domains, including travel, cooking, academia, apple, and aviation. We adopt WikiHowQA as the source dataset for transfer learning due to its high quality and large quantity, while StackExchange are used as the target dataset.

\subsection{Implementation Details}
We train all the implemented models with pre-trained GloVE embeddings\footnote{http://nlp.stanford.edu/data/glove.6B.zip} of 100 dimensions as word embeddings and set the vocabulary size to 50k for both source and target text. During training and testing procedure, we truncate the article to 400 words and restrict the length of generated summaries within 100 words. We apply early stopping based on the answer selection evaluation result on the validation set. We train our model and implement answer selection models for 5 epochs, while we implement summarization models for 20 epochs for fair comparisons, since the answers may repetitively occur in the candidates for different questions in the WikiHowQA dataset. 

In our model, we train with a learning rate of 0.15 and an initial accumulator value of 0.1. The dropout rate is set to 0.5. The hidden unit sizes of the BiLSTM encoder and the LSTM decoder are all set to 150. We train our models with the batch size of 32. All other parameters are randomly initialized from [-0.05, 0.05]. $\lambda_1$, $\lambda_2$, $\lambda_3$ are all set to 1.

\section{Experimental Result}

\subsection{Answer Selection Result}

We first compare the proposed method with several state-of-the-art methods on the answer selection task, including Siamese BiLSTM~\cite{DBLP:conf/aaai/MuellerT16}, Att-BiLSTM~\cite{Tan2016Improved}, AP-LSTM~\cite{dos2016attentive}, CA (Compare-Aggregate)~\cite{compare-aggregate} and COALA~\cite{COALA}. Besides, we perform several \textbf{Two-Stage} methods, which first summarize the original answers and then conduct answer selection. To validate the effectiveness of different components of ASAS, we also conduct ablation tests. MAP and MRR are adopted as evaluation metrics.

\begin{table}
\fontsize{8}{10}\selectfont
\centering
  \begin{tabular}{ccc}
    \toprule
    \textbf{Models} &\textbf{\textsc{MAP}}&\textbf{\textsc{MRR}}\\
    \midrule
    \textsc{Random Guess}&0.4088&0.4319\\
    \textsc{BM25}&0.4212&0.4377\\
    \textsc{Siamese BiLSTM}&0.4604&0.4734\\
    \textsc{Att-BiLSTM}&0.4573&0.4721\\
    \textsc{AP-BiLSTM}&0.4896&0.5058\\
    \textsc{CA}&\underline{0.5022}&\underline{0.5214}\\
    \textsc{COALA}&0.5003&0.5196\\
    \midrule
    GOLD + AP-BiLSTM&0.5261&0.5377\\
    PGN + AP-BiLSTM&0.4992&0.5078\\
    QPGN + AP-BiLSTM&0.5237&0.5343\\
    QPGN + CA&0.5246&0.5373\\
    QPGN + COALA&0.5197&0.5302\\
    \midrule
    \textbf{Joint Learning (ASAS)}&\textbf{0.5522}&\textbf{0.5686}\\
    w/o two-way attention&0.5208&0.5311\\
    w/o pointer network&0.5341&0.5483\\
  \bottomrule
\end{tabular}
\caption{Evaluation on Answer Selection}
\label{asresult}
\end{table}

Answer selection results on WikiHowQA are summarized in Table~\ref{asresult}. We show that the joint learning model (ASAS) achieves state-of-the-art performance. There are several notable observations in the results. (i) BM25 model and even the basic deep learning model slightly improve the performance compared to random guessing, which signifies that the testing set is indeed a difficult one. (ii) The Compare-Aggregate methods (including CA and COALA) and AP-BiLSTM, which have been proven to be relatively effective in long-sentence answer selection~\cite{COALA,dos2016attentive}, outperforms other strong baseline methods. (iii) Although \textbf{Two-Stage} methods actually improve the final answer selection result, it is time-consuming and inconvenient to train two separate models. In specific, using gold summary (GOLD) achieves the best performance, and Question-driven PGN (QPGN) performs better than original PGN. With the same summarization method, different answer selection models achieve similar results. (iv) Finally, the proposed joint learning model (ASAS) decently and substantially enhances the performance, which not only achieves the SOTA result, but also is easily trained by end-to-end fashion. By doing so, we precisely pick out the correct answers from candidate answers with long sentences, and meanwhile generate abstractive summaries for the convenience of community users. (v) The ablation study shows both the two-way attention mechanism and the pointer network contribute to the final result. The two-way attention mechanism enhances the interaction between questions and decoded answer summaries, while the pointer network aids in generating a better summary.

\subsection{Answer Summary Generation Result}

To evaluate the generated answer summary, we also compare the proposed method with the following state-of-the-art baseline methods on text summarization subtask, including four extractive methods (Lead3, TextRank~\cite{DBLP:conf/emnlp/MihalceaT04}, NeuralSum~\cite{DBLP:conf/acl/0001L16}, NeuSum~\cite{DBLP:conf/acl/ZhaoZWYHZ18}), two abstractive methods (Seq2Seq~\cite{DBLP:conf/conll/NallapatiZSGX16}, PGN~\cite{DBLP:conf/acl/SeeLM17}) and two query-based methods ($\text{SD}_2$~\cite{DBLP:conf/acl/NemaKLR17}, biASBLSTM~\cite{DBLP:conf/ecir/SinghMOBK18}). ROUGE F1 scores are used to evaluate the summarization methods.

\begin{table}
\fontsize{8}{10}\selectfont
\centering
  \begin{tabular}{cccc}
    \toprule
    \textbf{Models}& \textbf{\textsc{ROUGE 1}} &\textbf{\textsc{ROUGE 2}}&\textbf{\textsc{ROUGE L}}\\
    \midrule
    \textsc{Lead3}&24.66&5.56&22.67\\
    \textsc{TextRank}&26.42&7.12&23.79\\
    \textsc{NeuralSum}&\underline{27.01}&6.78&25.10\\
    \textsc{NeuSum}&26.78&6.88&25.14\\
    \textsc{Seq2Seq w/ Attention}&20.31&5.53&19.75\\
    \textsc{PGN w/ coverage}&26.83&\underline{7.54}&\underline{25.20}\\
    \textsc{$\text{SD}_2$}&26.65&6.92&24.77\\
    \textsc{biASBLSTM}&24.74&6.02&22.75\\
    \midrule
    \textbf{Question-driven PGN}&27.32&7.98&25.46\\
    \textbf{Joint Learning (ASAS)}&\textbf{27.78}&\textbf{8.16}&\textbf{25.86}\\
  \bottomrule
\end{tabular}
\caption{Evaluation on Text Summarization}
\label{comparisons}
\end{table}

Text summarization results on WikiHowQA are summarized in Table~\ref{comparisons}. The experimental results show that the question-driven PGN outperforms all the state-of-the-art methods of both extractive and abstractive summarization, which demonstrates the effectiveness of incorporating question information to generate summaries for answers. The question information directly involves in the calculation of the generation probability to determine the next word whether generated from the vocabulary or copied from the source text. In addition, jointly learning with answer selection, ASAS further improves the result with a noticeable margin. The correlation information between question-answer pairs also aids in attending important words in the original answer, which are related to the question. These results show that ASAS can effectively generate high-quality summaries for the selected answers.

\subsection{Analysis of The Length of Answers}
\begin{figure}
\centering
\includegraphics[width=0.35\textwidth]{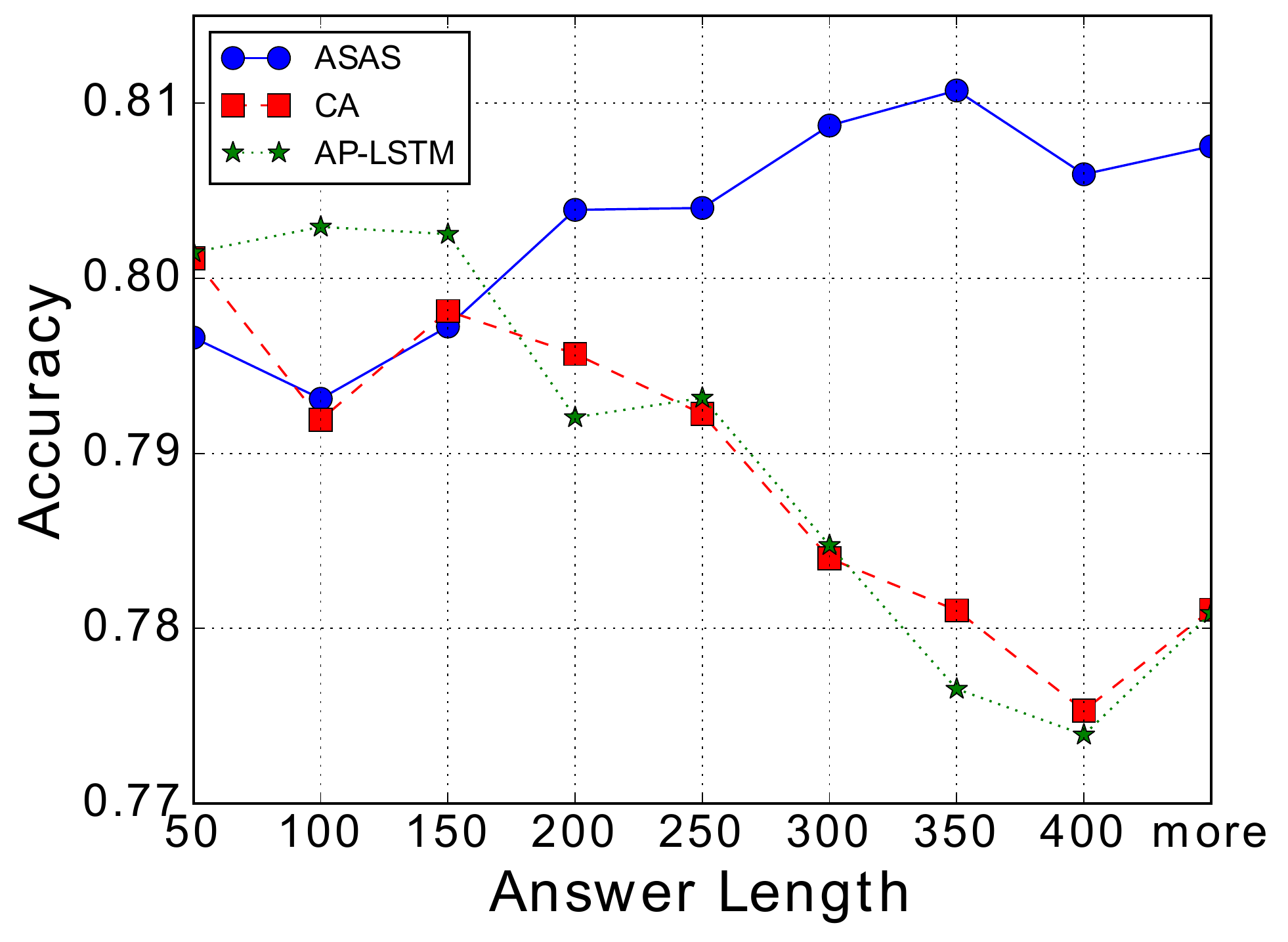}
\caption{Model Accuracy in terms of Answer Length}
\label{length}
\end{figure}

In order to validate the effectiveness of the proposed method on long-sentence answer selection, we split the test set in terms of the length of the answer. As shown in Fig.~\ref{length}, we compare ASAS with two baseline methods, AP-LSTM and Compare-Aggregate Model (CA), by measuring the accuracy, which is the ratio of correctly selected answers. We observe that ASAS performs better especially for long answers. For answers that are shorter than 100 words, CA and AP-LSTM is slightly better than ASAS, which indicates that the summary may have lost some information for short answers. However, the performance of these two methods goes down with the increase in the answer length, while ASAS maintains a great stability.

\begin{table}
\fontsize{8}{10}\selectfont
\centering
  \begin{tabular}{ccccc}
  \toprule
  Method & Info & Conc & Read & Corr\\
  \midrule
  \textsc{NeuralSum} & 3.60&2.70&3.22&3.24\\
  \textsc{PGN w/ coverage} &2.90&3.51&3.09&3.04\\
  \textbf{\textsc{ASAS}} & \textbf{3.67}&\textbf{3.88}&\textbf{3.59}&\textbf{3.71}\\
  \bottomrule
  \end{tabular}
\caption{Human Evaluation Results}
\label{human_eval}
\end{table}

\subsection{Human Evaluation on Summarization}
We conduct human evaluation on a sample of test set to evaluate the generated answer summaries from four aspects: (1) Informativity: how well does the summary capture the key information from the original answer? (2) Conciseness: how concise the summary is? (3) Readability: how fluent and coherent the summary is? (4) Correlatedness: how correlated the summary and the given question are? We randomly sample 50 answers and generate their summaries by three methods, including NeuralSum, PGN w/ coverage and the proposed ASAS. Three data annotators are asked to score each generated summary with 1 to 5 (higher the better).

\begin{figure}
\centering
\includegraphics[width=0.48\textwidth]{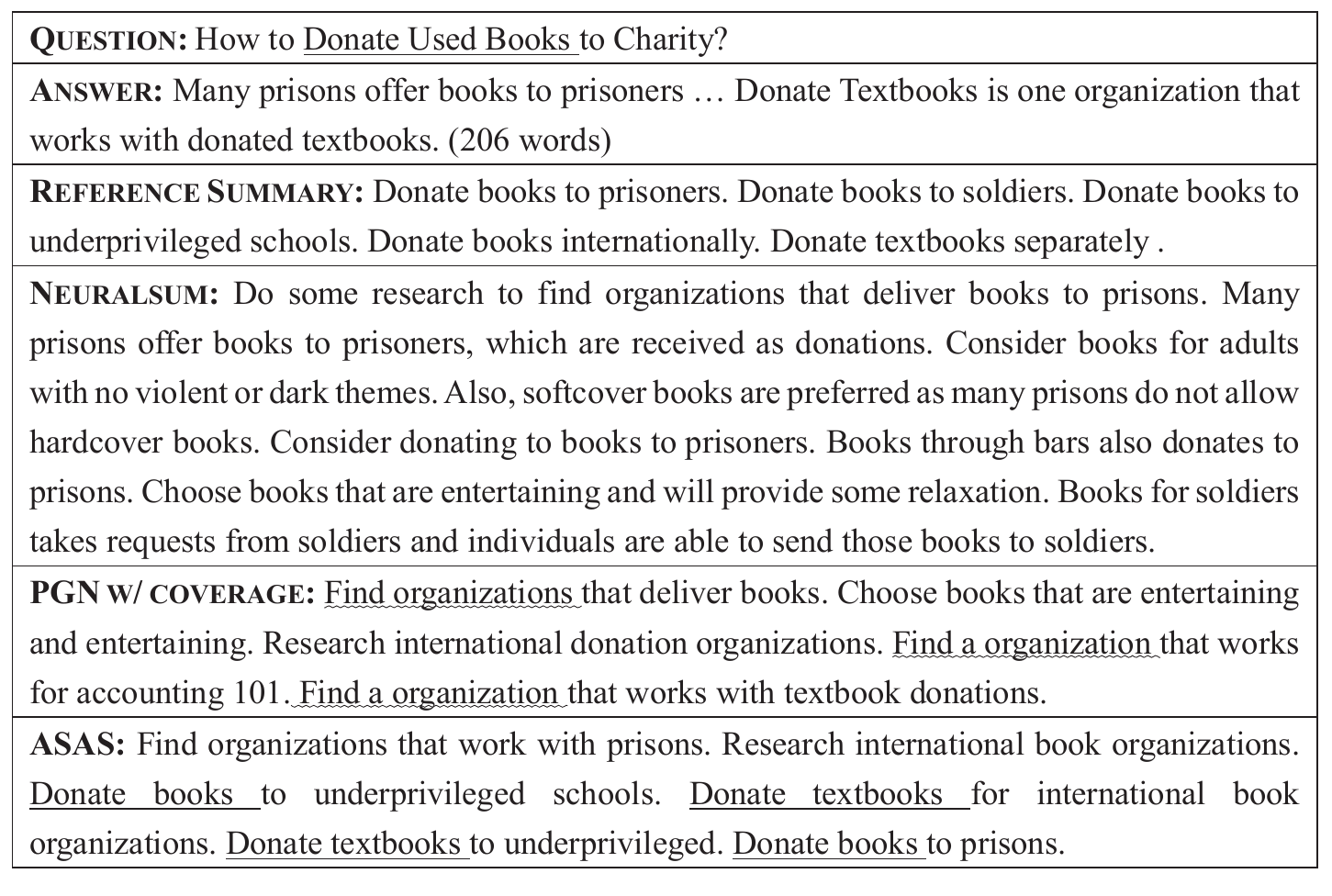}
\caption{Case Study. ASAS generates the answer summary highly related to the question (Underlined), while PGN may misunderstand the core idea of the answer (Wavy-lined).}
\label{case}
\end{figure}

Table~\ref{human_eval} shows the human evaluation results. The results show that ASAS consistently outperforms other methods in all aspects. Noticeably, the proposed method learns well to generate answer summaries that are highly related to the given questions so there is a substantial margin on Correlatedness. In order to intuitively observe the advantage of the proposed method, we randomly choose one example to show the answer summary generation results.  As shown in the Fig.~\ref{case}, the extractive method (e.g., NeuralSum) selects important sentences from the original answer to form the answer summary, which still contains many insignificant or redundant information. The abstractive method (e.g., PGN) generates the answer summary from the vocabulary and the original answer, which may miss some key words and essential information. Upon these defects, the proposed joint learning method (ASAS) takes into account the information provided by the question to capture the core idea of the original answer and generate a precise summary. More importantly, unlike other methods, answer summaries are generated at the same time that the answers are selected.

\subsection{Resource-poor CQA Results}
To evaluate the transferring ability and applicability of the proposed method, we conduct experiments on the resource-poor CQA task with transfer learning. We also conduct several ablations that use no pre-training or no fine-tuning, including (i) \textit{Finetune/-} is the baseline without pre-training, (ii) \textit{Finetune/No} is trained with the training set of source data without fine-tuning on the target training data, (iii) \textit{Finetune/Yes} is to first pre-train a model on the source data, and then use the learned parameters to initialize the model parameters for only fine-tuning the answer selection part on the target data. Following previous studies~\cite{COALA}, we adopt the ratio of correctly selected answers as the evaluation metrics. Note that we use an unsupervised summarization method, TextRank~\cite{DBLP:conf/emnlp/MihalceaT04}, to generate reference summaries roughly for \textit{Finetune/-} settings with ASAS, since there is no reference summary in the original StackExchange dataset.

\begin{table}
\fontsize{6}{7}\selectfont
\centering
  \begin{tabular}{ccccccc}
    \toprule
    \textbf{Models}&Finetune&Travel&Cooking&Academia&Apple&Aviation\\
    \midrule
    \textsc{BM25}&-&38.1& 30.9& 29.2& 21.8& 37.0\\
    \textsc{BiLSTM}&-&45.3& 35.2& 31.5& 27.2& 37.3\\
    \textsc{Att-BiLSTM}&-&43.0 &36.2 &31.2 &24.7& 33.9\\
    \textsc{AP-BiLSTM}&-&38.8& 32.2 &27.3 &22.9& 34.5\\
    \textsc{CA}&-& 46.5& 39.4& 36.1 &29.2& 46.5\\
    \textsc{COALA} &-&\underline{53.8}& 47.3& \underline{42.2}& 32.0& 48.4\\
    \midrule
    AP-BiLSTM&No&39.7&34.4&30.6&25.7&34.8\\
    CA&No&33.4&28.1&21.4&21.2&31.5\\
    COALA&No&35.6&32.2&24.5&22.8&37.2\\
    AP-BiLSTM&Yes&44.9&38.1&36.7&29.1&46.3\\
    CA&Yes&46.2&39.9&36.6&29.5&45.2\\
    COALA&Yes&52.7&\underline{49.2}&41.5&\underline{32.4}&\underline{49.9}\\
    \midrule
    \textbf{ASAS}&-&54.8&48.1&42.8&32.6&50.1\\
    \textbf{ASAS}&No&52.3&45.8&39.9&30.9&48.2\\
    \textbf{ASAS}&Yes&\textbf{56.5}&\textbf{52.8}&\textbf{44.4}&\textbf{35.1}&\textbf{52.9}\\
  \bottomrule
\end{tabular}
\caption{Evaluation on Resource-poor Answer Selection}
\label{transfer}
\end{table}

The experimental results show that even with the coarse reference summaries, ASAS (Finetune/-) achieves the best performance in 4 out of 5 domains, which demonstrates the applicability of the proposed joint learning framework. Under the zero-shot setting, ASAS (Finetune/No) also achieves competitive results as those strong baseline methods, which shows the strong transferring ability of the proposed method and the value of the large-scale source dataset, WikiHowQA. Fine-tuning the answer selection part further outperforms all the baselines by about 4\%.  This result indicates that there are actually some gaps between different CQA datasets and the fine-tune strategy effectively overcomes these domain differences. Compared with ASAS and AP-BiLSTM, CA and COALA hardly benefit from pre-training due to their reliance on unsupervised embedding matching features. 

\begin{figure}
\centering
\includegraphics[width=0.48\textwidth]{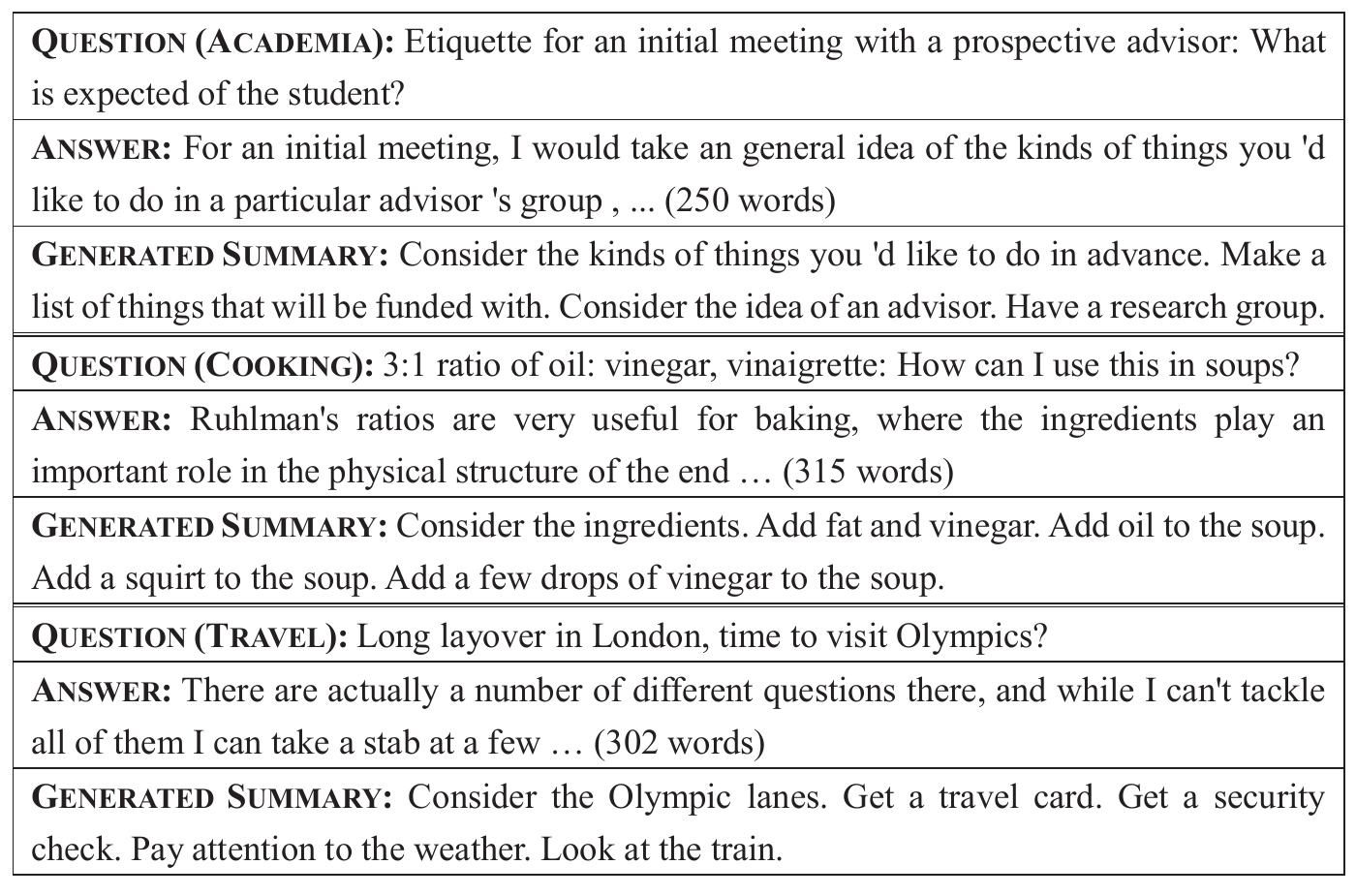}
\caption{Generated Summaries for Resource-poor CQA}
\label{case2}
\end{figure}

In addition, Fig.~\ref{case2} presents examples of answer summary generation results from target datasets. For those resource-poor CQA tasks without reference answer summaries, ASAS can not only achieve state-of-the-art results on answer selection, but also automatically generate decent and concise summaries via a simple transfer learning strategy with a resource-rich dataset. 

\section{Conclusion}
We study the joint learning of answer selection and answer summary generation in CQA. We propose a novel model to employ the question information to improve the summarization result, and meanwhile leverage the summaries to reduce noise in answers for a better performance on long-sentence answer selection. 
In order to evaluate the answer generation task in CQA, we construct a new large-scale CQA dataset, WikiHowQA, which contains both labels for answer selection task and reference summaries for text summarization task. The experimental results show that the proposed joint learning method outperforms the state-of-the-art methods on both answer selection and summarization tasks, and processes robust applicability and transferring ability for resource-poor CQA tasks. 

\bibliographystyle{aaai}
\bibliography{AAAI-DengY.5372.bib}

\begin{thebibliography}{}

\bibitem[\protect\citeauthoryear{Cheng and
  Lapata}{2016}]{DBLP:conf/acl/0001L16}
Cheng, J., and Lapata, M.
\newblock 2016.
\newblock Neural summarization by extracting sentences and words.
\newblock In {\em ACL}.

\bibitem[\protect\citeauthoryear{Choi \bgroup et al\mbox.\egroup
  }{2017}]{DBLP:conf/acl/ChoiHUPLB17}
Choi, E.; Hewlett, D.; Uszkoreit, J.; Polosukhin, I.; Lacoste, A.; and Berant,
  J.
\newblock 2017.
\newblock Coarse-to-fine question answering for long documents.
\newblock In {\em ACL},  209--220.

\bibitem[\protect\citeauthoryear{Cohen, Yang, and
  Croft}{2018}]{DBLP:conf/sigir/CohenYC18}
Cohen, D.; Yang, L.; and Croft, W.~B.
\newblock 2018.
\newblock Wikipassageqa: {A} benchmark collection for research on non-factoid
  answer passage retrieval.
\newblock In {\em SIGIR},  1165--1168.

\bibitem[\protect\citeauthoryear{Deng \bgroup et al\mbox.\egroup
  }{2018}]{DBLP:conf/coling/DengSYLDFL18}
Deng, Y.; Shen, Y.; Yang, M.; Li, Y.; Du, N.; Fan, W.; and Lei, K.
\newblock 2018.
\newblock Knowledge as {A} bridge: Improving cross-domain answer selection with
  external knowledge.
\newblock In {\em COLING},  3295--3305.

\bibitem[\protect\citeauthoryear{Deng \bgroup et al\mbox.\egroup
  }{2019}]{DBLP:conf/aaai/DengXLYDFLS19}
Deng, Y.; Xie, Y.; Li, Y.; Yang, M.; Du, N.; Fan, W.; Lei, K.; and Shen, Y.
\newblock 2019.
\newblock Multi-task learning with multi-view attention for answer selection
  and knowledge base question answering.
\newblock In {\em AAAI},  6318--6325.

\bibitem[\protect\citeauthoryear{dos Santos \bgroup et al\mbox.\egroup
  }{2016}]{dos2016attentive}
dos Santos, C.~N.; Tan, M.; Xiang, B.; and Zhou, B.
\newblock 2016.
\newblock Attentive pooling networks.
\newblock {\em CoRR, abs/1602.03609}.

\bibitem[\protect\citeauthoryear{Joty, M{\`{a}}rquez, and
  Nakov}{2018}]{DBLP:conf/emnlp/JotyMN18}
Joty, S.~R.; M{\`{a}}rquez, L.; and Nakov, P.
\newblock 2018.
\newblock Joint multitask learning for community question answering using
  task-specific embeddings.
\newblock In {\em EMNLP},  4196--4207.

\bibitem[\protect\citeauthoryear{Koupaee and
  Wang}{2018}]{DBLP:journals/corr/abs-1810-09305}
Koupaee, M., and Wang, W.~Y.
\newblock 2018.
\newblock Wikihow: {A} large scale text summarization dataset.
\newblock {\em CoRR} abs/1810.09305.

\bibitem[\protect\citeauthoryear{Li \bgroup et al\mbox.\egroup
  }{2017}]{DBLP:conf/wsdm/LiDLXFLGS17}
Li, Y.; Du, N.; Liu, C.; Xie, Y.; Fan, W.; Li, Q.; Gao, J.; and Sun, H.
\newblock 2017.
\newblock Reliable medical diagnosis from crowdsourcing: Discover trustworthy
  answers from non-experts.
\newblock In {\em WSDM},  253--261.

\bibitem[\protect\citeauthoryear{Mihalcea and
  Tarau}{2004}]{DBLP:conf/emnlp/MihalceaT04}
Mihalcea, R., and Tarau, P.
\newblock 2004.
\newblock Textrank: Bringing order into text.
\newblock In {\em EMNLP},  404--411.

\bibitem[\protect\citeauthoryear{Moschitti, Bonadiman, and
  Uva}{2017}]{DBLP:conf/eacl/MoschittiBU17}
Moschitti, A.; Bonadiman, D.; and Uva, A.
\newblock 2017.
\newblock Effective shared representations with multitask learning for
  community question answering.
\newblock In {\em EACL},  726--732.

\bibitem[\protect\citeauthoryear{Mueller and
  Thyagarajan}{2016}]{DBLP:conf/aaai/MuellerT16}
Mueller, J., and Thyagarajan, A.
\newblock 2016.
\newblock Siamese recurrent architectures for learning sentence similarity.
\newblock In {\em AAAI}.

\bibitem[\protect\citeauthoryear{Nakov \bgroup et al\mbox.\egroup
  }{2017}]{DBLP:conf/semeval/NakovHMMMBV17}
Nakov, P.; Hoogeveen, D.; M{\`{a}}rquez, L.; Moschitti, A.; Mubarak, H.;
  Baldwin, T.; and Verspoor, K.
\newblock 2017.
\newblock Semeval-2017 task 3: Community question answering.
\newblock In {\em SemEval@ACL},  27--48.

\bibitem[\protect\citeauthoryear{Nallapati \bgroup et al\mbox.\egroup
  }{2016}]{DBLP:conf/conll/NallapatiZSGX16}
Nallapati, R.; Zhou, B.; dos Santos, C.~N.; G{\"{u}}l{\c{c}}ehre, {\c{C}}.; and
  Xiang, B.
\newblock 2016.
\newblock Abstractive text summarization using sequence-to-sequence rnns and
  beyond.
\newblock In {\em CoNLL}.

\bibitem[\protect\citeauthoryear{Nallapati, Zhai, and
  Zhou}{2017}]{DBLP:conf/aaai/NallapatiZZ17}
Nallapati, R.; Zhai, F.; and Zhou, B.
\newblock 2017.
\newblock Summarunner: {A} recurrent neural network based sequence model for
  extractive summarization of documents.
\newblock In {\em AAAI},  3075--3081.

\bibitem[\protect\citeauthoryear{Nema \bgroup et al\mbox.\egroup
  }{2017}]{DBLP:conf/acl/NemaKLR17}
Nema, P.; Khapra, M.~M.; Laha, A.; and Ravindran, B.
\newblock 2017.
\newblock Diversity driven attention model for query-based abstractive
  summarization.
\newblock In {\em ACL},  1063--1072.

\bibitem[\protect\citeauthoryear{Nishida \bgroup et al\mbox.\egroup
  }{2019}]{DBLP:conf/acl/NishidaSNSOAT19}
Nishida, K.; Saito, I.; Nishida, K.; Shinoda, K.; Otsuka, A.; Asano, H.; and
  Tomita, J.
\newblock 2019.
\newblock Multi-style generative reading comprehension.
\newblock In {\em ACL},  2273--2284.

\bibitem[\protect\citeauthoryear{R{\"{u}}ckl{\'{e}}, Moosavi, and
  Gurevych}{2019}]{COALA}
R{\"{u}}ckl{\'{e}}, A.; Moosavi, N.~S.; and Gurevych, I.
\newblock 2019.
\newblock {COALA:} {A} neural coverage-based approach for long answer selection
  with small data.
\newblock In {\em AAAI},  6932--6939.

\bibitem[\protect\citeauthoryear{Saha \bgroup et al\mbox.\egroup
  }{2018}]{DBLP:conf/acl/KhapraSSA18}
Saha, A.; Aralikatte, R.; Khapra, M.~M.; and Sankaranarayanan, K.
\newblock 2018.
\newblock Duorc: Towards complex language understanding with paraphrased
  reading comprehension.
\newblock In {\em ACL},  1683--1693.

\bibitem[\protect\citeauthoryear{See, Liu, and
  Manning}{2017}]{DBLP:conf/acl/SeeLM17}
See, A.; Liu, P.~J.; and Manning, C.~D.
\newblock 2017.
\newblock Get to the point: Summarization with pointer-generator networks.
\newblock In {\em ACL},  1073--1083.

\bibitem[\protect\citeauthoryear{Severyn and
  Moschitti}{2015}]{Severyn2015Learning}
Severyn, A., and Moschitti, A.
\newblock 2015.
\newblock Learning to rank short text pairs with convolutional deep neural
  networks.
\newblock In {\em SIGIR},  373--382.

\bibitem[\protect\citeauthoryear{Shen \bgroup et al\mbox.\egroup
  }{2018}]{DBLP:conf/sigir/ShenDYLD0L18}
Shen, Y.; Deng, Y.; Yang, M.; Li, Y.; Du, N.; Fan, W.; and Lei, K.
\newblock 2018.
\newblock Knowledge-aware attentive neural network for ranking question answer
  pairs.
\newblock In {\em SIGIR},  901--904.

\bibitem[\protect\citeauthoryear{Singh \bgroup et al\mbox.\egroup
  }{2018}]{DBLP:conf/ecir/SinghMOBK18}
Singh, M.; Mishra, A.; Oualil, Y.; Berberich, K.; and Klakow, D.
\newblock 2018.
\newblock Long-span language models for query-focused unsupervised extractive
  text summarization.
\newblock In {\em ECIR},  657--664.

\bibitem[\protect\citeauthoryear{Song \bgroup et al\mbox.\egroup
  }{2017}]{DBLP:conf/wsdm/SongRLLMR17}
Song, H.; Ren, Z.; Liang, S.; Li, P.; Ma, J.; and de~Rijke, M.
\newblock 2017.
\newblock Summarizing answers in non-factoid community question-answering.
\newblock In {\em WSDM},  405--414.

\bibitem[\protect\citeauthoryear{Tan \bgroup et al\mbox.\egroup
  }{2016}]{Tan2016Improved}
Tan, M.; Santos, C.~D.; Xiang, B.; and Zhou, B.
\newblock 2016.
\newblock Improved representation learning for question answer matching.
\newblock In {\em ACL},  464--473.

\bibitem[\protect\citeauthoryear{Tomasoni and
  Huang}{2010}]{DBLP:conf/acl/TomasoniH10}
Tomasoni, M., and Huang, M.
\newblock 2010.
\newblock Metadata-aware measures for answer summarization in community
  question answering.
\newblock In {\em ACL},  760--769.

\bibitem[\protect\citeauthoryear{Wang and Jiang}{2017}]{compare-aggregate}
Wang, S., and Jiang, J.
\newblock 2017.
\newblock A compare-aggregate model for matching text sequences.
\newblock In {\em ICLR}.

\bibitem[\protect\citeauthoryear{Wang and
  Manning}{2010}]{DBLP:conf/coling/WangM10a}
Wang, M., and Manning, C.~D.
\newblock 2010.
\newblock Probabilistic tree-edit models with structured latent variables for
  textual entailment and question answering.
\newblock In {\em COLING},  1164--1172.

\bibitem[\protect\citeauthoryear{Wang and Nyberg}{2015}]{DBLP:conf/acl/WangN15}
Wang, D., and Nyberg, E.
\newblock 2015.
\newblock A long short-term memory model for answer sentence selection in
  question answering.
\newblock In {\em ACL},  707--712.

\bibitem[\protect\citeauthoryear{Wang \bgroup et al\mbox.\egroup
  }{2018a}]{DBLP:conf/aaai/WangYGWKZCTZJ18}
Wang, S.; Yu, M.; Guo, X.; Wang, Z.; Klinger, T.; Zhang, W.; Chang, S.;
  Tesauro, G.; Zhou, B.; and Jiang, J.
\newblock 2018a.
\newblock R\({}^{\mbox{3}}\): Reinforced ranker-reader for open-domain question
  answering.
\newblock In {\em AAAI},  5981--5988.

\bibitem[\protect\citeauthoryear{Wang \bgroup et al\mbox.\egroup
  }{2018b}]{DBLP:conf/acl/XiaoWLWL18}
Wang, Z.; Liu, J.; Xiao, X.; Lyu, Y.; and Wu, T.
\newblock 2018b.
\newblock Joint training of candidate extraction and answer selection for
  reading comprehension.
\newblock In {\em ACL},  1715--1724.

\bibitem[\protect\citeauthoryear{Wang, Ming, and
  Chua}{2009}]{DBLP:conf/sigir/WangMC09}
Wang, K.; Ming, Z.; and Chua, T.
\newblock 2009.
\newblock A syntactic tree matching approach to finding similar questions in
  community-based qa services.
\newblock In {\em SIGIR},  187--194.

\bibitem[\protect\citeauthoryear{Wen \bgroup et al\mbox.\egroup
  }{2018}]{DBLP:conf/aaai/WenMFZ18}
Wen, J.; Ma, J.; Feng, Y.; and Zhong, M.
\newblock 2018.
\newblock Hybrid attentive answer selection in {CQA} with deep users modelling.
\newblock In {\em AAAI},  2556--2563.

\bibitem[\protect\citeauthoryear{Wu, Sun, and
  Wang}{2018}]{DBLP:conf/acl/WuWS18}
Wu, W.; Sun, X.; and Wang, H.
\newblock 2018.
\newblock Question condensing networks for answer selection in community
  question answering.
\newblock In {\em ACL},  1746--1755.

\bibitem[\protect\citeauthoryear{Yang \bgroup et al\mbox.\egroup
  }{2019}]{DBLP:conf/ijcai/0007CCWZS19}
Yang, M.; Chen, L.; Chen, X.; Wu, Q.; Zhou, W.; and Shen, Y.
\newblock 2019.
\newblock Knowledge-enhanced hierarchical attention for community question
  answering with multi-task and adaptive learning.
\newblock In {\em IJCAI},  5349--5355.

\bibitem[\protect\citeauthoryear{Yang, Yih, and Meek}{2015}]{Yang2015WikiQA}
Yang, Y.; Yih, W.~T.; and Meek, C.
\newblock 2015.
\newblock Wikiqa: A challenge dataset for open-domain question answering.
\newblock In {\em EMNLP},  2013--2018.

\bibitem[\protect\citeauthoryear{Yoon, Shin, and
  Jung}{2018}]{DBLP:conf/naacl/YoonSJ18}
Yoon, S.; Shin, J.; and Jung, K.
\newblock 2018.
\newblock Learning to rank question-answer pairs using hierarchical recurrent
  encoder with latent topic clustering.
\newblock In {\em NAACL-HLT},  1575--1584.

\bibitem[\protect\citeauthoryear{Yu \bgroup et al\mbox.\egroup
  }{2018}]{DBLP:conf/wsdm/YuQJHSCC18}
Yu, J.; Qiu, M.; Jiang, J.; Huang, J.; Song, S.; Chu, W.; and Chen, H.
\newblock 2018.
\newblock Modelling domain relationships for transfer learning on
  retrieval-based question answering systems in e-commerce.
\newblock In {\em WSDM},  682--690.

\bibitem[\protect\citeauthoryear{Zhou \bgroup et al\mbox.\egroup
  }{2018}]{DBLP:conf/acl/ZhaoZWYHZ18}
Zhou, Q.; Yang, N.; Wei, F.; Huang, S.; Zhou, M.; and Zhao, T.
\newblock 2018.
\newblock Neural document summarization by jointly learning to score and select
  sentences.
\newblock In {\em ACL},  654--663.

\bibitem[\protect\citeauthoryear{Zhou, Lin, and
  Hovy}{2006}]{DBLP:conf/lrec/ZhouLH06}
Zhou, L.; Lin, C.; and Hovy, E.~H.
\newblock 2006.
\newblock Summarizing answers for complicated questions.
\newblock In {\em LREC},  737--740.

\end{thebibliography}

\end{document}